\documentclass[letterpaper, 10 pt, conference]{ieeeconf}
\IEEEoverridecommandlockouts
\usepackage{amsmath}
\usepackage{amssymb}
\usepackage{amsfonts}
\let\labelindent\relax
\usepackage{enumitem}
\usepackage{graphicx}
\usepackage{float}
\usepackage{textcomp}

\newcommand{\citep}[1]{\cite{#1}}

\title{\LARGE \bf
``The wallpaper is ugly'': \\ Indoor Localization using Vision and Language}

\author{Seth Pate and Lawson L.S. Wong
\thanks{$^{1}$Khoury College of Computer Sciences,
        Northeastern University,
        Boston, MA\;\;
        {\tt\small pate.s@northeastern.edu ; lsw@ccs.neu.edu}}%
}

\begin{document}

\maketitle

\begin{abstract}
    We study the task of locating a user in a mapped indoor environment using natural language queries and images from the environment.
    Building on recent pretrained vision-language models, we learn a similarity score between text descriptions and images of locations in the environment.
    This score allows us to identify locations that best match the language query, estimating the user's location.
    Our approach is capable of localizing on environments, text, and images that were not seen during training.
    One model, finetuned CLIP, outperformed humans in our evaluation.
\end{abstract}

\section{Introduction}

Natural language is an important medium of communication between humans and robots~\cite{Tellex2020}.
Many robot tasks refer to, and rely on understanding, the robot's spatial environment.
Connecting natural language with the robot's spatial knowledge is therefore critical.
However, making this connection is challenging because humans tend to represent and express knowledge about the environment differently from robots.
Natural language descriptions tend to be high-level, sparse, and semantic, whereas robot spatial representations are low-level, dense, and geometric.

\begin{figure}
\begin{center}
\includegraphics[width=.85\linewidth]{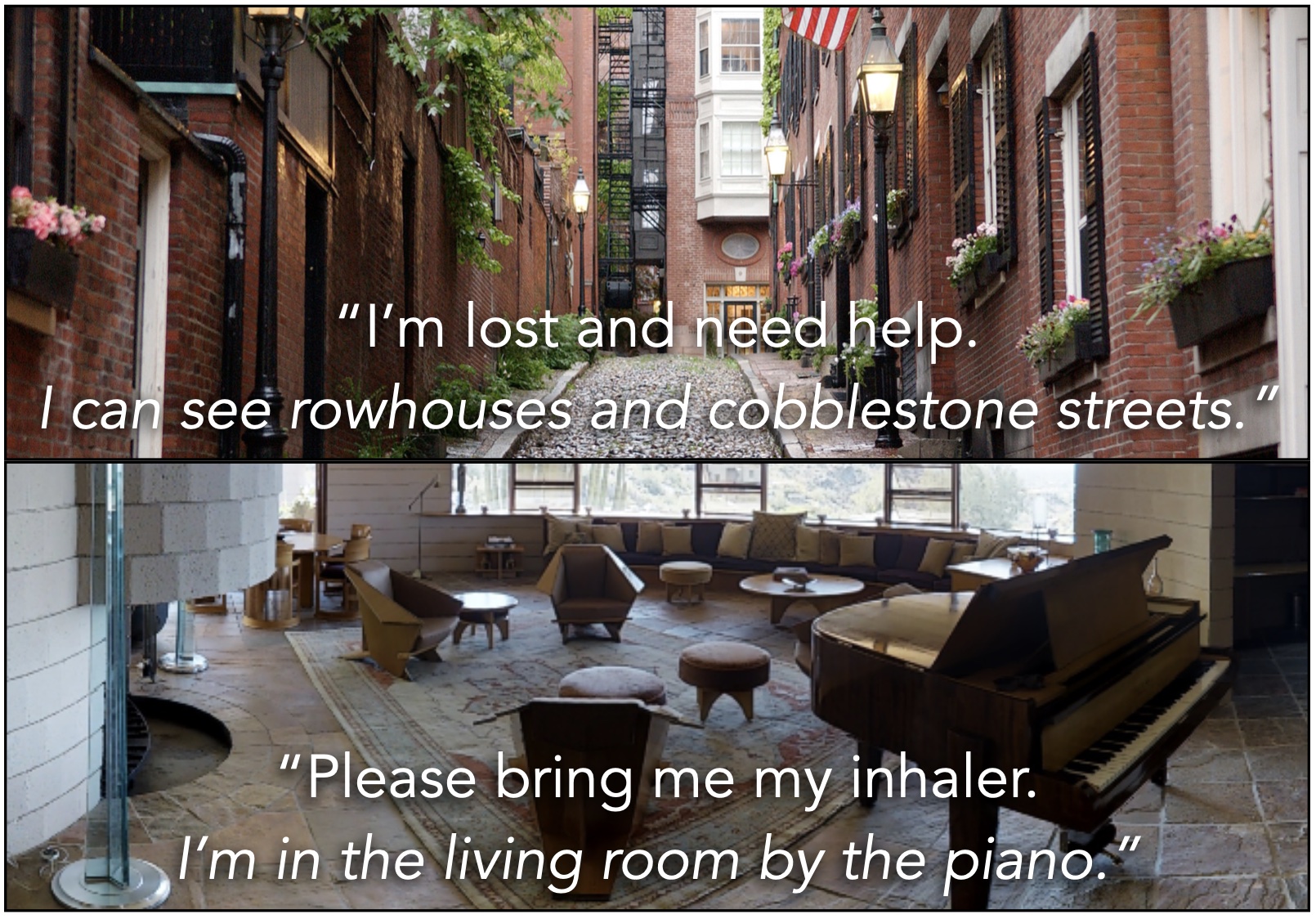}
\caption{\textbf{Use of localization in robotics.}
    Localization is a necessary first step when a robot must help a human without perfect knowledge of their location. This may apply to search and rescue (top) or household assistance (bottom). In this paper, we study only the localization task. Photo credits: Ian Howard (top), Matterport3D (bottom).}
\label{fig:cartoon}
\end{center}
\end{figure}

\begin{figure}[h!]
\begin{center}
\includegraphics[width=.85\linewidth]{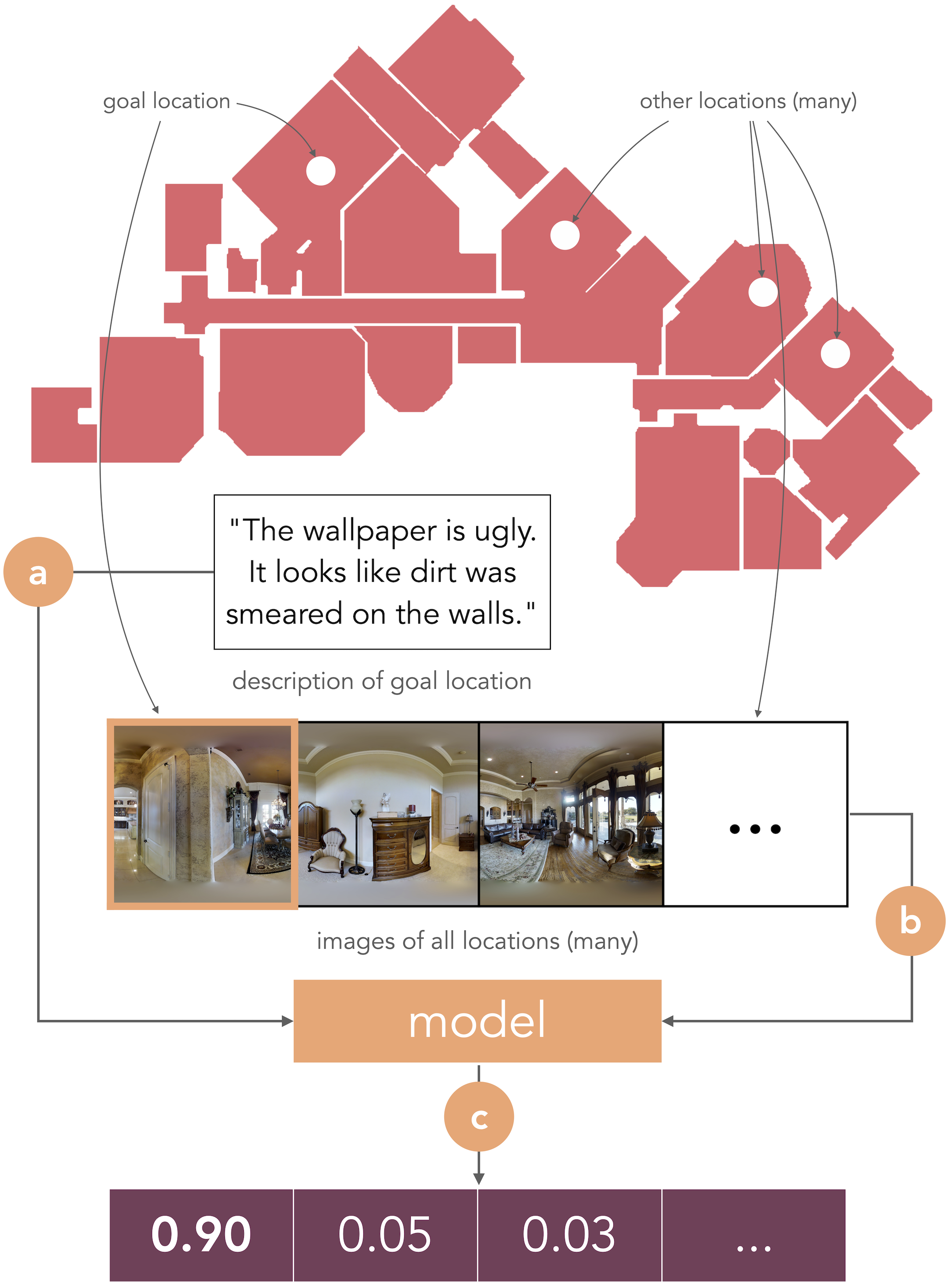}
\caption{\textbf{Vision-language localization.}
    (a) The model encodes the user's description of their location, the goal.
    (b) The model encodes an exhaustive sample of images representing all locations in the environment.
    (c) The model produces a similarity score between each image and the description, which, after softmax, outputs a distribution to predict the user's location.}
\label{fig:intro}
\end{center}
\end{figure}

In this paper, we propose and study one important problem in this intersection:
locating a user in a known environment, given a natural language description of a desired location.
This is an important capability in several robot applications, such as finding someone who is lost by asking them to describe their surroundings, or autonomous delivery, where the robot needs to communicate with and find the recipient via natural language.
Location is also the first step in providing spatial directions to a user -- before you can tell someone where to go, you need to know where they are.

\begin{figure*}[t]
\begin{center}
\includegraphics[width=.75\linewidth]{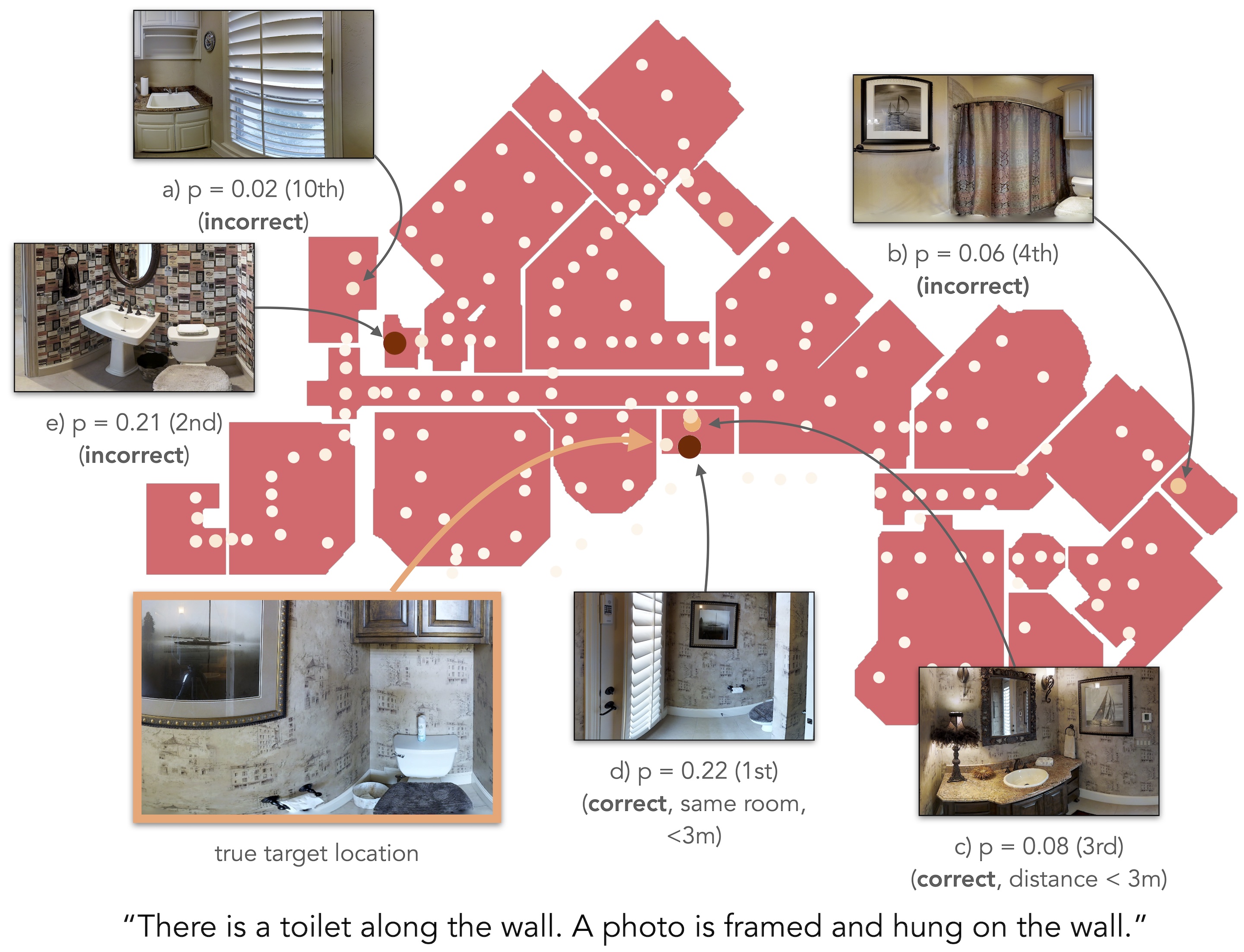}
\caption{\textbf{Example Model Output.} Our model creates a likelihood distribution across the 170 locations in this \textit{scan}. The model’s confidence is shown by both the size and color of the circles, which represent \textit{views}. We highlight some guesses alongside the true target location, a bathroom. From top left, clockwise:
(a) The 10th guess is a laundry room with a sink, but no toilet.
(b) The 4th guess has a photo and a toilet. It is a good guess, but the wrong bathroom.
(c) The 3rd guess was taken from the hallway, but has a clear view of the target location.
(d) The model’s best guess is in the same annotated region (room), adjacent to the target.
(e) The 2nd guess is in another bathroom without a framed photo, only a mirror which may resemble one.}
\label{fig:modeloutput}
\end{center}
\end{figure*}

We illustrate our task and high-level approach in Fig.~\ref{fig:intro}.
We assume that the robot has already mapped the environment and knows all the locations that may be queried.
We use Matterport3D~\citep{Matterport3D} as a source of mapped indoor environments with rich visual information.
The user, located at some point in the environment, describes their surroundings in natural language.
With this sentence and images from a discrete set of locations in the mapped environment,
we compute a learned similarity score between the description query and each location's images.
We can use the resulting distribution of scores to estimate the user's location.

To relate descriptions to images, we use a large pretrained vision-language model~\citep{Du2022}, CLIP~\citep{radford2021learning}, which learns complex representations of images and text on different pretraining tasks where data is widely available.

To improve the model's performance, we repurposed finetuning data from datasets that are related to our task.
Specifically, we used data for vision and language navigation~\citep{anderson2018vision,rxr,Wang_2022_CVPR,Gu2022}
to construct two additional datasets for our task (`RxR' and `RxR\_landmarks'), which gave significant improvements.

To evaluate the model, we gathered a small test dataset of human descriptions for locations in Matterport3D.
Then we performed two experiments. 
First, we compared different finetuning subsets, as well as an alternate model using convolutional neural nets (CNNs) and a long short-term memory (LSTM) text encoder.
Then we compared the best finetuned model to a human baseline on the test set. 
In this setting, the model outperformed the human baseline. 

In summary, we define the task of vision-language localization, consider a simple approach using existing pretrained vision-language models,
collect and construct several datasets to finetune the pretrained model, and evaluate our models on localizing in Matterport3D indoor environments.

\section{Related Work}
\label{sec:related}

Natural language is widely used in robotics; see Tellex \textit{et~al.}~\citep{Tellex2020} for a comprehensive survey. 
As the survey indicates, the primary focus of previous work has been on instruction following and answering/asking questions.
`Vision and language navigation'~\citep{anderson2018vision} (VLN) extends the instruction following concept directly to
sighted agents moving within indoor or outdoor environments~\citep{hong2021vln, krantz2020beyond, chen2021history, krantz2023iterative, chen2019touchdown}. A lot of VLN work, including ours, is done with the Matterport3D enviroment, which we describe below.~\citep{Matterport3D}

Although many VLN papers ask embodied agents to navigate a certain path between two points, others simply ask agents to find an object or location within an environment, which is very similar to our location task. 
Examples include REVERIE~\cite{qi2020reverie} and SOON~\cite{zhu2021soon}, whose benchmarks use deep recurrent neural networks, reinforcement learning, imitation learning, and graph embeddings to approach the problem.

Whereas these papers treat location as a navigation problem and ask agents to produce a series of actions, we simply ask agents to rank target locations in a known environment. 
This is more similar to Hahn~\textit{et al.}~\citep{hahn2020you} and Chen~\textit{et al.}~\citep{chen2020scanrefer}, who encode language instructions alongside a given environment to produce, respectively, a distribution of locations or a bounding box in that environment.

Our approach is different, in that we take environments as a set of unrelated images, and treat location as an image retrieval problem. 
This allows us to use readily available pretrained image similarity models which we can finetune.

We describe our finetuning datasets below in sec. \ref{sec:data}. We opted not to include several related datasets from the works cited above: the From Anywhere to Object (FAO)~\citep{zhu2021soon} is more concerned with specific objects than locations, and the Where Are You? (WAY)~\citep{hahn2020you} dataset uses dialogue rather than our choice of a single utterance. The relatively small size of these datasets (in the thousands of samples, on the order of our test set) suggests they may not be large enough to affect finetuning.

\section{Matterport3D Environment}

Matterport3D is a collection of RGB-D images taken of indoor spaces
by a Matterport panoramic camera.
The dataset has 90 \textit{scans} of buildings, mostly elaborate homes
and a few oddities like cruise ship cabins and spas.
Each \textit{scan} is divided into a navigable graph of \textit{viewpoints}
or \textit{views}, which are spread throughout at the house at a spacing of $\sim$2.2m.
Finally, each \textit{view} contains 36 RGB-D images, which can be knit together
into a panoramic.
We use equirectangular panoramics provided by Rey-Area~\textit{et~al.}~\citep{bata1126},
although they were only able to reliably create images for about 85\% of
the \textit{views} in the dataset.
This limited our choice of data to those \textit{views} 
we had coverage for (see Fig.~\ref{fig:dataoverlap}).

\section{Task and Architecture}

We define our task in the Matterport3D environment.
A user occupies one of $M$ \textit{views}, $v_m$, in a \textit{scan}, $s$.
The user gives a description, $d$, of their \textit{view} to the agent.
The agent has an image $i_m$ for each $v \in s$,
and produces a distribution
$$
P(v_m \,|\, d, i_1, \ldots, i_M) \ \ \forall\ \ v\ \in\ s \quad (1 \leq m \leq M)
$$
The agent may then guess which \textit{view} the user occupies.
A sample of our model's output is shown by Fig.~\ref{fig:modeloutput}.

Defined this way, the problem is an \textit{image-text similarity} task,
suitable for large, pretrained transformer networks.
OpenAI's Contrastive Language-Image Pretraining (CLIP)~\citep{radford2021learning}, popular and readily available for finetuning, is a useful tool in vision and language tasks~\cite{shen2021much}.
Recently, simple CLIP models have done very well in benchmarks like the RoboTHOR and Habitat ObjectNav challenges~\cite{khandelwal2022simple}.
We used the \texttt{vit-base-patch32} model, the smaller vision transformer variant~\cite{HuggingFaceCLIP}.

\begin{figure}[t]
\begin{center}
\includegraphics[width=.9\linewidth]{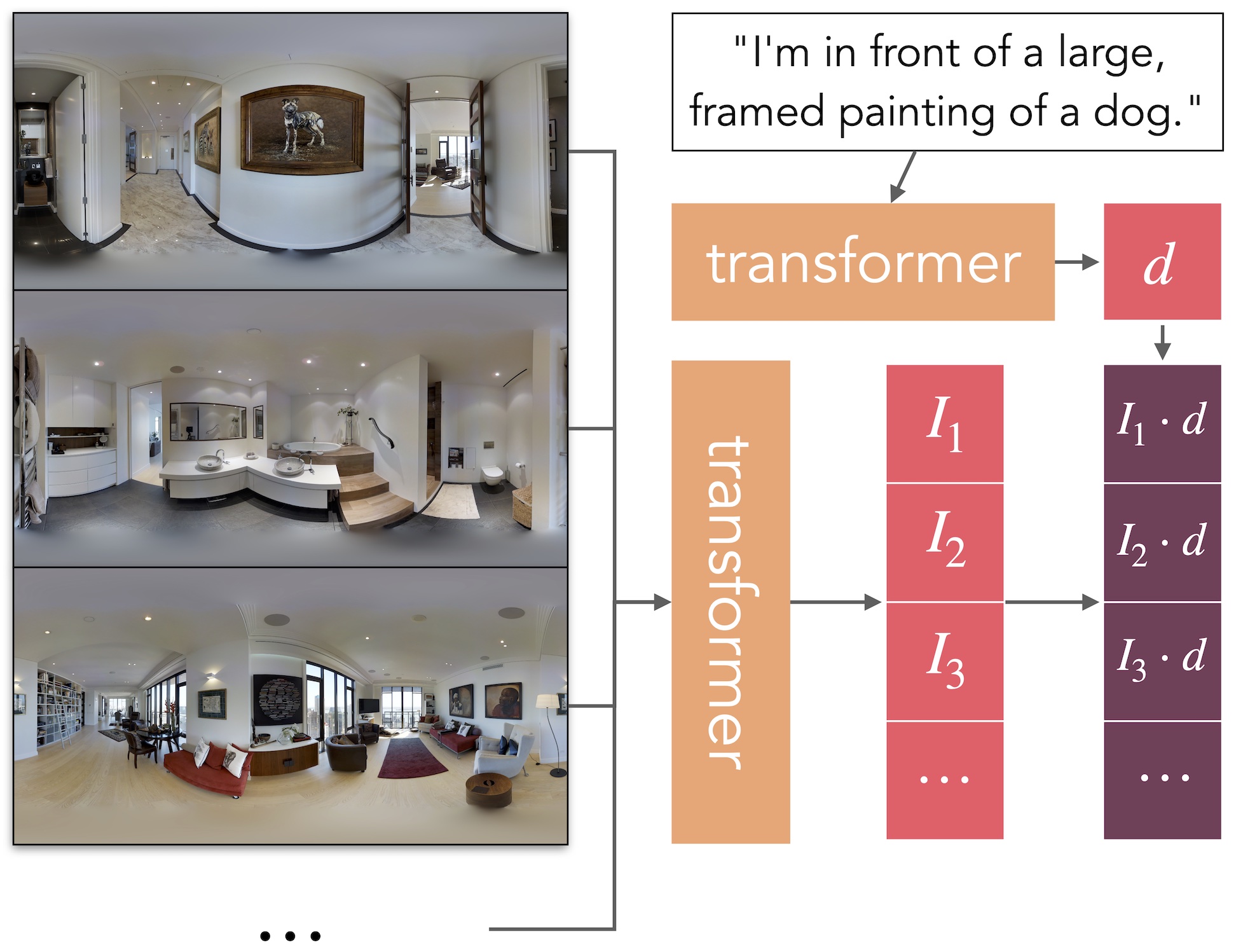}
\caption{\textbf{Model.} CLIP~\citep{radford2021learning} uses transformer networks~\citep{Vaswani2017, dosovitskiy2020image} to encode text and images into vectors of identical length, then compares these vectors by taking their dot product. In our task, a description might be compared with as many as 170 images (\textit{views}) from the environment (\textit{scan}).}
\label{fig:model}
\end{center}
\end{figure}

CLIP encodes $N$ texts and $M$ images with separately trained encoders.
The dot product of the encodings are returned as $(N,M)$ logit similarity scores (see Fig.~\ref{fig:model}).

To train CLIP, we encode a batch of $N$ (text, image) pairs. 
A `perfect' model produces a square $(N,N)$ matrix which, after applying
softmax, becomes the identity matrix.
We use this matrix as a target for binary cross entropy loss and
optimize the CLIP network with gradient descent.
We used an Adam optimizer~\citep{kingma2014adam} with learning rate $5 \times 10^{-7}$ and weight decay $10^{-3}$.
Finetuning took less than a day on an RTX 2080 Ti.

CLIP was pretrained on a large ($10^8$) dataset called WebImageText (WIT)~\citep{radford2021learning}.
On our human test set, described below, pretrained CLIP performs well above random as a zero-shot classifier.
We use pretrained CLIP as a baseline measure in our experiments,
and improve its performance through finetuning, described below.

\section{Data}
\label{sec:data}

\begin{figure}[t]
\begin{center}
\includegraphics[width=.9\linewidth]{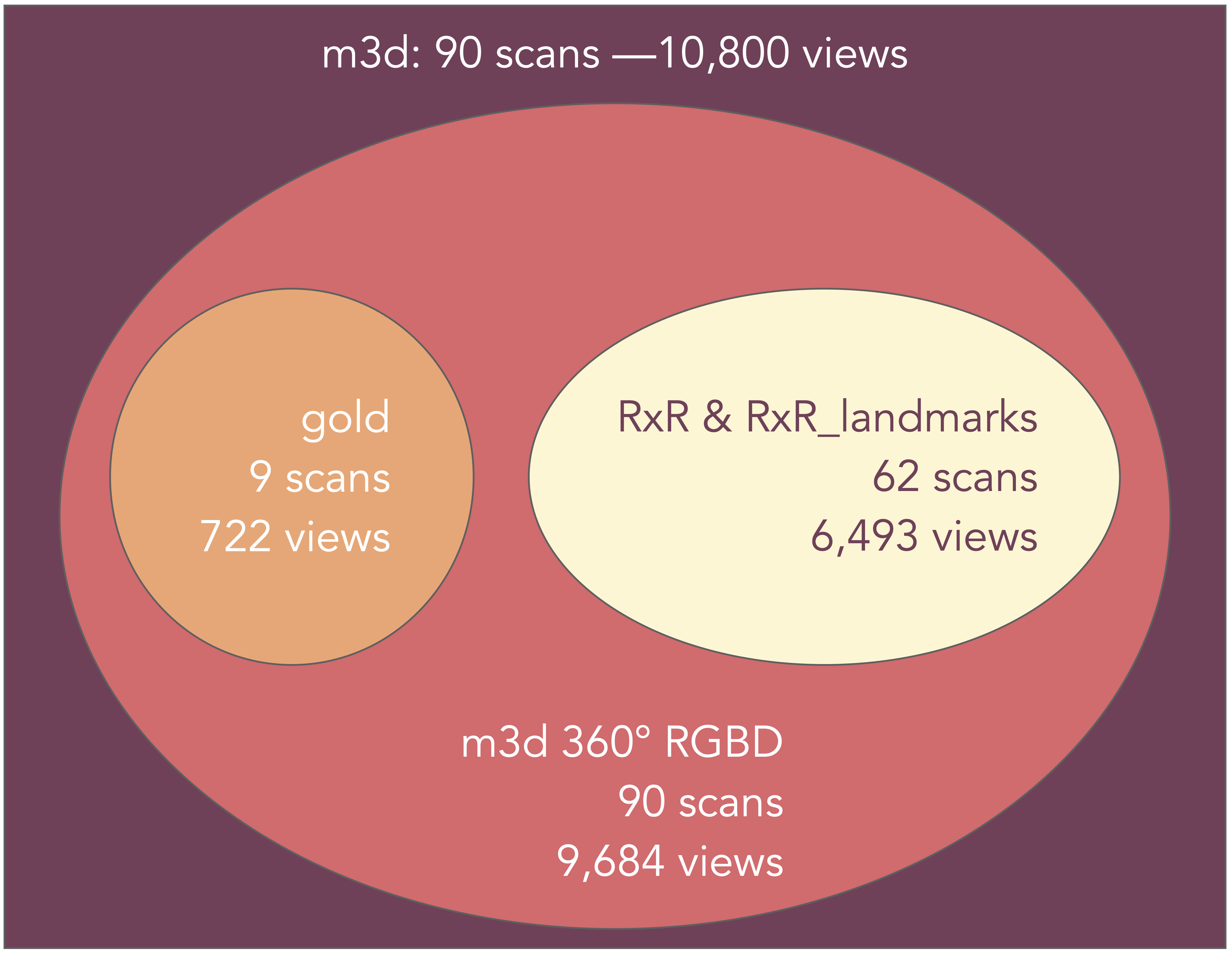}
\caption{(a) The Matterport3D (m3d) 360\textdegree\, RGBD set (red) contains most, but not all, of the images in Matterport3D (purple). We used the former for its equirectangular format.
(b) The human `gold’ test set (orange) is disjoint from the finetuning sets RxR and RxR\_landmarks (beige). Both are subsets of the m3d 360\textdegree\, RGBD set.}
\label{fig:dataoverlap}
\end{center}
\end{figure}

We used three datasets: a human test set which we collected,
    and the `RxR' and `RxR\_landmarks' datasets,
    which we repurposed from existing data and used to finetune our model (see Table~\ref{tab:data} and Fig.~\ref{fig:dataoverlap}).

\begin{table}[t]
  \caption{Data Statistics (`m3d' refers to Matterport3D)}
  \label{tab:data}
  \begin{center}
    \begin{tabular}{l|r|r|r|r}
        set & \# scans & \# views & \# samples & avg \# words \\
        \hline
        gold & 9 & 722 & 1,443 & 23 \\
        RxR & 62 & 6,493 & 205,092 & 14 \\
        RxR\_landmarks & 62 & 6,472 & 172,309 & 2 \\
        m3d 360\textdegree\, RGBD & 90 & 9,684 & n/a & n/a \\
        m3d & 90 & 10,800 & n/a & n/a \\
    \end{tabular}
    \end{center}
\end{table}

\begin{figure}[t]
\begin{center}
\includegraphics[width=.8\linewidth]{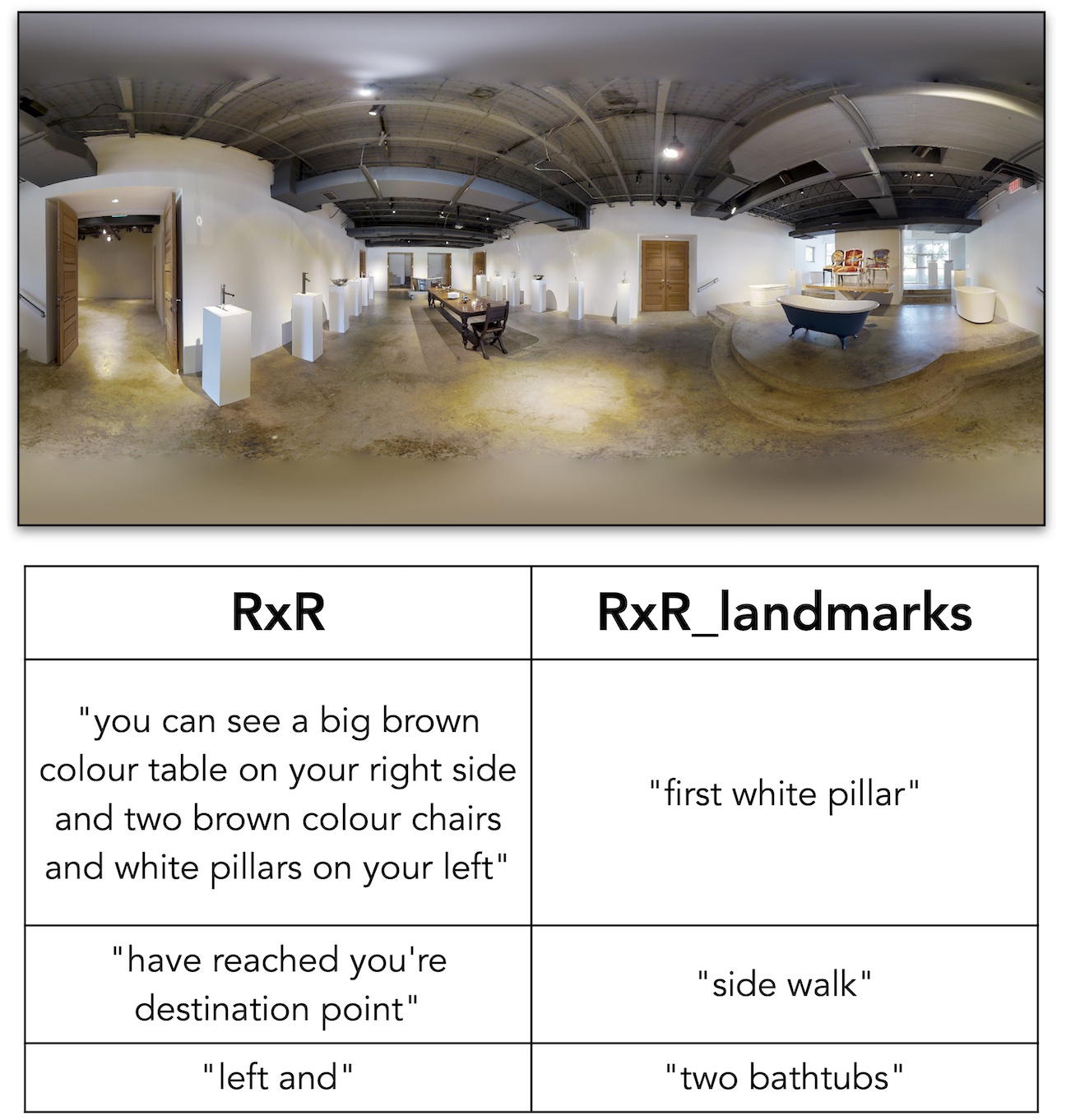}
\caption{We drew from two existing datasets to create a finetuning set for CLIP. RxR, an instruction dataset, generally provided longer examples with more extensive grammar, but many fragments were unrelated to the image. RxR\_landmarks, on the other hand, provided very relevant, but very short, keyword samples.}
\label{fig:dataexamples}
\end{center}
\end{figure}

\begin{figure}[t!]
\begin{center}
\includegraphics[width=.9\linewidth]{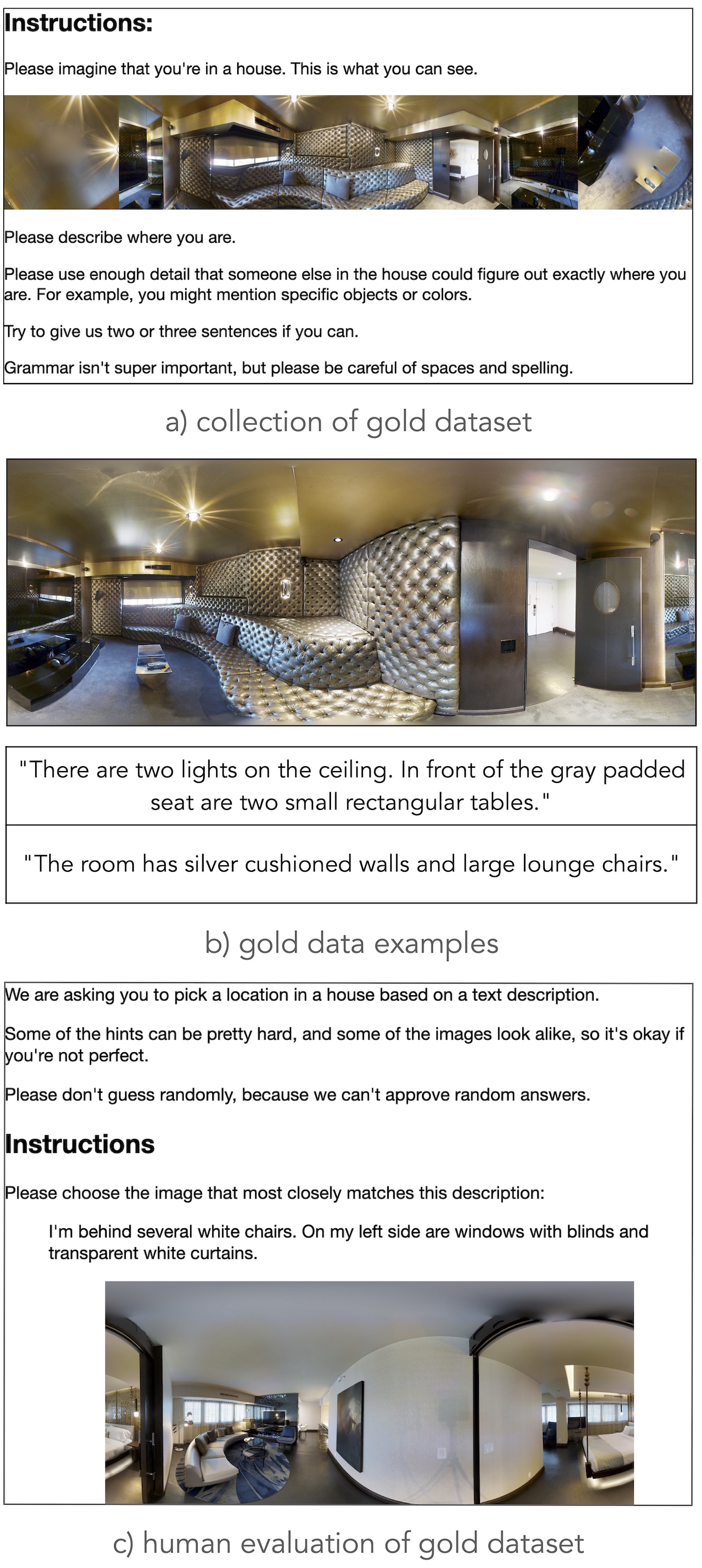}
\caption{We collected an evaluation dataset by asking users of Amazon Mechanical Turk (AMT) to describe their surroundings when shown panoramas from the Matterport3D dataset. (a) Our `gold' dataset has two samples for each \textit{view} in nine buildings, about 10\% of the total Matterport3D set. (b) We evaluated each sample by asking another AMT worker to pick the correct image from a collection of 20. (c) Each `gold’ sample was evaluated once.}
\label{fig:datacollection}
\end{center}
\end{figure}

\subsection{Human (`gold') set}

To evaluate our model, we needed human descriptions of locations in Matterport3D.
We chose 9 representative \textit{scans} from the 90 total in Matterport3D,
with 722 \textit{views} between them.
This covers about 10\% of the Matterport3D environment.
We collected two descriptions per \textit{view}, for 1,443 samples total.
Different humans wrote each description.

We used Amazon Mechanical Turk (AMT) to collect our data (see Fig.~\ref{fig:datacollection}).
For each sample, we show the worker a skybox image of a \textit{view} and ask them to describe their location,
so that another person who knew the space could find them.

\subsection{Room Across Room (`RxR')}

RxR~\citep{rxr} is a navigation dataset in the Matterport3D environment.
It contains, among other things, human descriptions of paths
in \textit{scans}. 
Each path is a sequence of \textit{views}.
Human `guides' were asked to describe their journey along this path
so that another user could follow it.

RxR has been used to evaluate language grounding models~\citep{rxr}, and most recently
to generate instructions for human evaluation~\cite{Wang_2022_CVPR}, but not for our particular
task of location. However, it is a high quality source of captioned
imagery specific to our environment, so we used it to finetune our model.

Samples in RxR contain a series of panoramic images (the path) and guide annotations describing the path.
Each word in the annotation is time stamped, 
allowing us to map it to a \textit{view} along the path,
giving us 205,092 (description, \textit{view}) pairs (see Fig.~\ref{fig:dataexamples}).

RxR includes image masks to indicate what parts of a \textit{view} were visible
to the guide when they said the word, but we omitted those masks here.

\begin{table*}
  \caption{Comparing Models Across Entire Gold Dataset}
  \label{tab:ablation}
  \begin{center}
    \begin{tabular}{l|l|r|rrr|r|r}
        model & data & success (\%) $\uparrow$ & hits at 1 (\%) $\uparrow$ & close (\%) $\uparrow$ & same room (\%) $\uparrow$ & error (m) $\downarrow$ & mrr $\uparrow$ \\
        \hline
        random & & 11 & 1 & 7 & 8 & 15.73 & 0.06 \\
        CNN-LSTM & rxr + landmarks & 11 & 1 & 7 & 9 & 16.68 & 0.06 \\
        CLIP & no pretrain, rxr + landmarks & 14 & 3 & 11 & 11 & 15.75 & 0.09 \\
        CLIP & pretrain only & 44 & 10 & 34 & 36 & 10.10 & 0.23 \\
        CLIP & pretrain + landmarks & 47 & 12 & 37 & 41 & 8.75 & 0.26 \\
        CLIP & pretrain + rxr & \textbf{55} & \textbf{14} & 43 & 46 & 7.74 & \textbf{0.28} \\
        CLIP & pretrain + rxr + landmarks & \textbf{55} & \textbf{14} & \textbf{44} & \textbf{47} & 
        \textbf{7.30} & \textbf{0.28} \\
    \end{tabular}
    \end{center}
\end{table*}

\subsection{Room Across Room `landmarks'}

Wang~\textit{et~al.}~\citep{Wang_2022_CVPR} developed the `landmarks' dataset from the original RxR data
to help generate instructions to guide users in the Matterport3D environment.
It contains 172,309 samples.
For each guide annotation of a path in RxR, Wang~\textit{et~al.} first used a transformer
to identity `entities' in the text, for example a couch, bathtub, or door.
They associated these entities with specific regions
(bounding boxes) of panoramic images along the path.

The resulting (landmark, region) pairs were helpful in generating instructions
to guide human users.
We use this data directly as our RxR\_landmarks dataset  (see Fig.~\ref{fig:dataexamples}).
We omit the bounding boxes, so that our data simply maps entities to entire
\textit{views}.

\section{Evaluation and Results}
\label{sec:results}

In this section, we describe our metrics for evaluating our model, and give the results of two experiments:
a comparison of different finetuning datasets and models on the full task, and 
then a comparison of our best model against a human baseline on a smaller, easier version of our task.

\subsection{Metrics}

\begin{enumerate}[leftmargin=*]
    \item \textbf{success} (\%): Percent success rate. The model is `successful' if its guess satisfies one of the following:
    \begin{itemize}[leftmargin=*]
        \item \textit{hits at 1} (\%): The guess is the target image.
        \item \textit{close} (\%): The guess is less than 3m from the target image, in graph distance. The average spacing in Matterport3D is $\sim$2.2m, so 3m is a common success threshold in related literature using this environment. All \textit{hits at 1} pass this test.
        \item \textit{same room} (\%): Matterport3D \textit{scans} are annotated by hand into `regions' like bathroom, living room, etc. We say that the model is successful if it was able to guess the correct region. All \textit{hits at 1} pass this test.
    \end{itemize}
    Fig.~\ref{fig:modeloutput} shows some of these success conditions.
    \item \textbf{error} (m): The average graph distance between the guess and the target \textit{view}.
    \item \textbf{mean reciprocal rank} (MRR): $\frac{1}{k}\sum_k\frac{1}{\texttt{rank}}$, where \texttt{rank} is the priority of the target in the model's confidence distribution. An MRR of $0.5$ indicates that the model is expected to rank the target image second.
\end{enumerate}

\subsection{Comparing Models and Finetuning}

In our first experiment, we compared different models on the original (all \textit{views}) task (see Table~\ref{tab:ablation}).
The random baseline chooses one of $M$ \textit{views} in the \textit{scan}.
The CNN-LSTM model encodes the \textit{view} image with a pretrained ResNet152~\cite{he2016deep} and the
description with a bidirectional, 3-layer LSTM encoder using the CLIP tokenizer.

The `no pretrain' model is a CLIP that is trained only on the combined finetuning data (no WIT),
whereas `pretrain only' is CLIP only trained on its WIT dataset.
Finally, we show some ablations of the finetuning dataset, comparing the relative
contribution of RxR and RxR\_landmarks to the combined dataset in the final row.

We show an example of our model's output in Fig.~\ref{fig:modeloutput}.
As the caption describes, the model is frequently able to identify key features
from the description.
But even when it does identify the right \textit{type} of room, it might fail to
pick the correct instance of that type -- a particular bathroom or bedroom, for example.

This experiment shows the importance of CLIP's pretraining.
The model without pretraining (Table~\ref{tab:ablation} row 3) performs little better than random (row 1).
Our finetuning dataset, at $10^5$ samples, is probably too small to train a large model like CLIP.
Similarly, the CNN-LSTM variant (row 2) performs only at random, despite the pretrained ResNet.

Finetuning CLIP (rows 5--7) increased the pretrained model's performance (row 4)
by up to 25\% on the test set.
This is a substantial gain, given that the finetuning set is three orders of magnitude smaller than the pretraining data.

We were surprised to see that the RxR\_landmarks set seems to add little beyond the RxR set.
The one or two word descriptions of landmarks are much shorter than our test set queries, which are
several sentences long.
Most of its landmark information is probably present in the RxR set.

\begin{table*}
  \caption{Comparison to Human Baseline (only 20 choices)}
  \label{tab:human}
  \begin{center}
    \begin{tabular}{l|l|r|rrr|r}
        model & data & success (\%) $\uparrow$ & hits at 1 (\%) $\uparrow$ & close (\%) $\uparrow$ & same room (\%) $\uparrow$ & error (m) $\downarrow$ \\
        \hline
        random & & 12 & 4 & 10 & 10 & 15.70 \\
        CLIP & pretrain only & 51 & 30 & 45 & 45 & 8.10 \\
        human & & 57 & \textbf{38} & 52 & 53 & 7.07 \\
        CLIP & pretrain + rxr + landmarks & \textbf{63} & \textbf{38} & \textbf{54} & \textbf{58} & \textbf{5.90}  \\
    \end{tabular}
    \end{center}
\end{table*}

\subsection{Human Baseline}

To test our model against a human baseline, we scaled down our task.
In our task defined above, the model compares the description to every \textit{view} in the
\textit{scan}.
A \textit{scan} might have well over a hundred \textit{views}.
Our model can process an arbitrary number of \textit{views}, but comparing that many similar images is very challenging for a human evaluator to do.

Instead, we presented workers on AMT with only 20 \textit{views},
randomly chosen from the same \textit{scan},
one of which is the true target \textit{view} which the query text describes.\footnote{The Northeastern University IRB has reviewed this research and exempted it from further action.}
We collected one evaluation for each sample in the \textit{gold} dataset (1,443),
and scored them according to the metrics above.
To control the quality of responses, we discarded data from workers who failed to score a `success' in 
at least 20\% of their responses, which is above random (12\%).

We compared our best finetuned CLIP model (Table~\ref{tab:human} row 4) to the human baseline (row 3). Like the humans, the model chose from only 20 images. In this experiment, the model slightly outperformed the human baseline.
However, we address some limitations of the human evaluation, and other aspects of our work, in the next section.

\section{Limitations, and Future Directions}
\label{sec:limitations}

\subsection{Human Evaluation}

We used AMT to get a human baseline on our task.
However, performance between workers varied significantly:
Some individuals scored a success rate of 60--70\%, others between 20--30\%.
This disparity was expected, given that humans vary in their ability to navigate and track spatial relationships.~\citep{kozhevnikov2006perspective}

An informal pilot study showed that graduate students scored about a 70\% success rate on the same task.
These students were very familiar with the dataset and the task,
so their performance might be a better estimate of good human performance.
However, we lacked the resources to evaluate the entire test set this way.

Both the informal and formal figures (70\% and 57\%) may seem low for a human evaluation.
We suggest two explanations.
First, humans were asked to choose from 20 independent images; this is a somewhat unnatural task and not how humans generally think of spatial environments.
Second, some descriptions of the environment are ambiguous and could easily refer to more than one location.
The example in fig. \ref{fig:modeloutput} shows that houses tend to have several bathrooms which all look quite similar.
Given these challenges, we think the human success rate is reasonable, and compares well to human performance on a similar location task in the same environment (70.4\% for human locators in WAY~\citep{hahn2020you}).

\subsection{Continuous Environment}

Human performance might increase with a better representation of our environment.
Our Matterport3D environment uses a navigable graph of \textit{views},
but more recent implementations use a continuous environment compatible with a simulator
like Habitat~\citep{krantz2020beyond, savva2019habitat}.

In a continuous environment, the user can move more naturally throughout a building,
exploring and stopping the simulation when they think they have found the right location.
This formulation of the task is more natural.
In addition to being easier for humans to understand, a continuous training environment
would help a robot prepare better for a real world demonstration.

\subsection{Improving Data}

A simpler improvement would be to use bounding box and mask information for our finetuning datasets.
As mentioned earlier, the RxR datasets (including RxR\_landmarks) have
image mask and bounding box data that might allow us to draw a tighter relationship
between text and image regions.
We omitted these bounding boxes, so that in the RxR\_landmarks set for example,
the word `refrigerator' is mapped to an entire panoramic image of a kitchen.
The model may perform better with a more specific mapping of concepts to pixels.

In addition to the bounding box information from the datasets we used, we expect that finetuned results would improve with additional relevant data, including those we described in section \ref{sec:related}.

\section{Conclusion}
\label{sec:conclusion}

We proposed the task of vision-language localization, and considered a simple approach using existing pretrained vision-language models.
We also collected and constructed several datasets for this task, to finetune the pretrained model.
We outperformed the baseline pretrained model,
as well as a human baseline on a simplified version of our task.
In future work, we plan to deploy this competitive approach on a robot operating in a continuous environment, to locate users using natural language.

\section{Acknowledgements}

This work was supported by NSF Grants \#2107256 and \#2142519.

\bibliography{ral}
\bibliographystyle{IEEEtran}

\end{document}